\documentclass{article}
\usepackage{spconf,amsmath,amsfonts,graphicx}

\title{ Quality Evaluation of Point Clouds: A Novel No-Reference Approach using Transformer-based Architecture}
\name{Author(s) Name(s)\thanks{Thanks to XYZ agency for funding.}}
\address{Author Affiliation(s)}

\name{Marouane Tliba$^{1}$, Aladine Chetouani$^{1}$, Giuseppe Valenzise$^{2}$ and Fréderic Dufaux$^{2}$}
\address{$^{1}$Laboratoire PRISME, Université d'Orléans, Orléans, France\\
$^{2}$Université Paris-Saclay, CNRS, CentraleSupélec, Laboratoire des signaux et systèmes
\\}

\begin{document}
%
\maketitle
\begin{abstract}
With the increased interest in immersive experiences, point cloud came to birth and was widely adopted as the first choice to represent 3D media. 
Besides several distortions that could affect the 3D content spanning from acquisition to rendering, efficient transmission of such volumetric content over traditional communication systems stands at the expense of the delivered perceptual quality. 
To estimate the magnitude of such degradation, employing quality metrics became an inevitable solution. In this work, we propose a novel deep-based no-reference quality metric that operates directly on the whole point cloud without requiring extensive pre-processing, enabling real-time evaluation over both transmission and rendering levels. To do so, we use a novel model design consisting primarily of cross and self-attention layers, in order to learn the best set of local semantic affinities while keeping the best combination of geometry and color information in multiple levels from basic features extraction to deep representation modeling. 
\end{abstract}
\section{Introduction}
\label{sec:intro}

Point clouds (PC) have become increasingly important in recent years due to their ability to capture detailed and accurate representations of real-world objects and environments. 
One of the key benefits of using point clouds is that they can capture a high level of detail, including small features and irregularities that may be difficult to capture with other 3D data formats. 
However, 3D point clouds can only accurately represent a 3D scene by incorporating a large amount of information, which can reach thousands or even millions of points. 
Therefore, lossy compression schemes are often used, and became an inevitable application. 
To optimize the development of immersive 3D experiences and achieve the best visual quality for a given bit-rate, accurate quality metrics are needed. As a result, the field of point cloud quality assessment (PCQA) has received significant attention from researchers in recent years. 

Point cloud perceptual quality can be assessed using subjective or objective methods. 
To this end, several effective Full reference objective approaches have been proposed.
These approaches can be classified into three main groups: Point-based, Feature-based and Projection-based metrics. Point-based metrics such as Point-to-Point (Po2Po) \cite{Po2Po}, Point-to-Plane (Po2Pl) \cite{Po2Pl}, Plane-to-Plane (Pl2Pl) \cite{Pl2Pl} and Point-to-Mesh (Po2Mesh) \cite{Po2PM}, predict the quality through point-wise geometric and/or features distance between the reference PC and its distorted version. Po2Po measures the relative distance between point pairs to estimate the final quality, Po2Pl extends Po2Po by projecting the error vector along the local normal, and Pl2Pl quantify the quality through measuring the angular similarity between surfaces associated to the points from the reference and degraded contents. Po2Mesh creates a polygonal mesh from the reference sample and then compute the distance between each distorted point and the corresponding surface. Currently, MPEG is adopting Po2Po MSE and Po2Pl MSE with the associated PSNR as the standard point cloud geometry quality metrics.
Feature-based PC quality metrics extract the geometry with the associated attributes from point-wise level in a global or local way. Among those metrics, we can cite PC-MSDM \cite{PC-MSDM} that extends the 2D SSIM metric \cite{SSIM} to PC  by considering local curvature statistics, the Geotex \cite{Geotex} metric that exploits the Local Binary Pattern (LBP) \cite{LBPSurvey} descriptors, and PCQM \cite{PCQM} that combines the geometry and color features.
In projection-based PC Quality Metrics, the 3D points or their associated features are projected into 2D regular grids \cite{AldineWithRef}.


Some No-reference approaches have been also proposed especially the ones based on deep learning. However, these methods still present notable limitations and disadvantages, for instance, in order to adapt the irregular structure of scattered 3D point clouds, it implies a heavy pre-processing 
as 2D images projection or transformation (voxelization). 
As a result, the introduced pre-processing turning also the No-reference quality assessment time consuming and empirically hard to achieve. 
Recently, new efficient deep-based metrics such as PointNet-SSNR 
 \cite{ICIPPOINTNET} and PointNet-DCCFR\cite{tliba2022point} have been released. They exploit the intrinsic features of point cloud data. 
 As a drawback, these methods focus on distinct portions of the point cloud, assuming that the perceived quality is the same over the whole point cloud.

Our proposed method for no-reference point cloud quality assessment addresses a significant limitation in previous approaches, which often lack the ability to effectively consider the complex structural relationships within a point cloud regions. By utilizing a multi-level self and cross-attention, we are able to capture the local semantic affinities of points, providing a more comprehensive representation for the evaluation of its visual quality.
To sum up, the main contributions of this paper are summarized as follows:

\begin{enumerate}

   \item We present a novel efficient end-to-end deep-based method for PCs quality assessment (PCQA) that operates directly on the whole point cloud, without the need for any projection or other transformation. This allows for a more empirical and comprehensive evaluation of the point cloud's quality.
   
   \item Our method is designed to capture the local semantic affinities in point representations and creates connectivity between distant features using self-attention. This facilitates better feature extraction from local regions.
   
   \item We propose a new system for processing geometry and color features in point clouds using a parallel two-stream architecture. The two-stream networks operate independently, while keep interacting with each other dynamically at multiple levels using a cross-attention network design.


\end{enumerate}

\section{Proposed Method}
\label{sec:method}

\begin{figure*}[t]
  \centering
  \centerline{\includegraphics[scale=0.450]{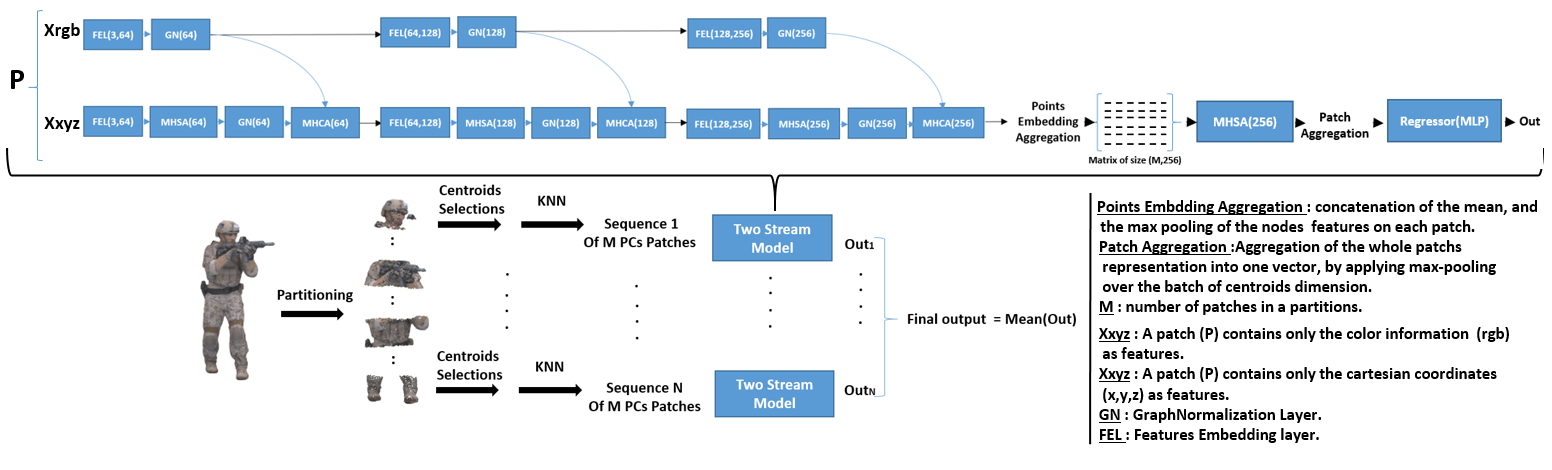}}

\caption{ General pipeline of our proposed method }
\label{fig:gen_arch}
 \vspace{-6mm}
\end{figure*}

As depicted in Fig. \ref{fig:gen_arch}, the overall pipeline of our method consists of three main steps:  starting with the pre-processing, then features extraction and representation mapping using self and cross-attention mechanisms, and finally quality estimation. 

\subsection{Pre-processing}

Pre-processing is a critical stage in our approach, aimed at optimizing the point cloud (PC) processing pipeline and consists of dividing the PC into vertical slices (partitions) of points. This partitioning has two primary objectives:  (i) enable parallel processing of points, and (ii) meet the memory requirements for GPU processing, as some PCs may have an important size (e.g millions of points). In order to balance the computation load, the number of partitions varies between 8 and 24 according to the size of the original PC. Each partition is then divided to form local patches. To achieve this, we first select a sufficient number of centroids and then apply the k-nearest neighbor clustering method.  We note that the selected distant centroids ensure to have an ensemble of patches that covers the whole partition' points. Each of the patches is then fed into our model for feature extraction and representation modeling. By applying this pre-processing method, we ensure that we provide our model  with a complete single enclosing all prominent features, and thus increase the model's ability to learn coherent local information from coherent regions.

\subsection{Proposed Transformer Model}
\subsubsection{Overview}

Our model design draws inspiration from two existing architectures: PointNet \cite{PointNetDL} and Transformer \cite{trans}. In particular, we employ a principal characteristic of PointNet in extracting information from point sets by using a permutation-invariant function. To do so, we used two parallel streams network for color and geometry independently. This network transforms the input point geometry and color information into a higher-dimensional space using a \textit{feature embedding} layer. To capture richer local structure information, we built on the \textit{feature embedding} of the geometry stream by introducing multi-head self-attention which draws the semantic affinities between neighboring points. 

In order to combine the geometry and the color representation, we employed  multi-head cross-attention. This results not only in adding the local connectivity information between adjacent point representations, but also completing it with corresponding color features. The resulting attentive connectivity of points' representation is updated dynamically at each level of the network, capturing different levels of semantic affinities, as the point embedding is updated. Our method thus combines the strengths of both PointNet and Transformer while introducing novel features for improved performance.

Overall, our proposed method shares almost the same philosophy as the original Transformer \cite{trans}. The PointNet feature embedding layers replace the input embedding and the subsequent feedforward networks. However, instead of capturing the context in an ordered sequence, our method is designed to find the semantic affinities between point representations without the need for an imposed order, as the input contains clear information about the geometry coordinates. 

\subsubsection{Two-stream Network}
We consider the geometry and color as independent  information, corresponding to two different processing streams. Depending on which stream we consider, input points $\mathbf{x}_i$ bring different information. For the geometry stream, $\mathbf{x}_i = (x_i, y_i, z_i)$ contains the 3-dimensional coordinates of the points, while for the color stream, points represent the RGB attribute information. Notice that it is possible to include additional features in other streams. 

Both the color and geometry streams consist of three \textit{feature embedding } layers, each of which  is comprised of a series of 1-D convolutions that are interspersed with nonlinear functions. Following each \textit{feature embedding } layer in the geometry stream, a multi-head self-attention layer is applied to draw the semantic affinities between neighboring points. Afterward, we employ cross-multi-head attention to align the learned color information with the geometrical representation.

Formally, for a given input point cloud' partition composed of $M$ patches $\{P^{0},P^{1},P^{2},..,P^{M}\}$, each 
 patch $P^{i}$ is represented as a matrix of size $\mathbb{R}^{N\times F}$, where $N$ is the number of points in the patch and $F$ is the number of features per point. On each layer, a stander \textit{feature embedding} layer $f_{\boldsymbol{\Theta}} : \mathbb{R}^F\rightarrow \mathbb{R}^{F'}$ is applied to produce a new representation of the provided ${X^{i}_{xyz}}$ and ${X^{i}_{rgb}}$ referring the geometry and color raw inputs or a subsequently produced representations, for each of the two streams.
 A multi-head self-attention layer ${MHSA_{\boldsymbol{\Theta}}} : \mathbb{R}^{F'}\rightarrow \mathbb{R}^{F'}$ is then applied to the geometry stream output ${f_{}}({X}_{xyz})$, producing an updated representation ${X'_{xyz}}$. Mathematically, the proposed ${MHSA_{\boldsymbol{\Theta}}}$  can be expressed as follows:
\begin{gather} 
\mathbf{q} = {X}_{xyz}W_q ,\mathbf{k} = {X}_{xyz}W_k,\mathbf{v} = {X}_{xyz}W_v \\
\mathbf{A} = \frac{softmax(\mathbf{q}\mathbf{k}^{T})}{\sqrt\frac{{C}}{h}}\\ {MHSA_{\boldsymbol{\Theta}}}({f_{}}({X}_{xyz})) = \mathbf{A}\mathbf{v}
\end{gather}
where ${W_q}, {W_k}, {W_v} \in {R}^{C×(C/h)} $
are learnable parameters, ${ C }$ and ${ h} $ are the embedding dimension and number of heads. 

The ${X'_{xyz}}$ representation can be considered as an embedding  induced from a graph propagation layer. As the  attention scores between the points create a sort of soft connection simulating the adjacency matrix. More precisely, the real connection between points could be obtained by setting an attention scores threshold. Therefore, to accelerate convergence, we incorporate a GraphNorm operation \cite{graphnorm} on both streams. Although the color representation ${X'_{rgb}}$ can not be deemed to have an origin from a graph-similar function, we find that applying the same sort of normalization on the two streams is useful to keep a fixed scale of the features. The GraphNorm operation is a variation of InstanceNorm \cite{Ulyanov2016InstanceNT}, tailored for graph normalization, and includes a learnable parameter $\alpha$ that determines how much of the channel-wise average to retain in the shift operation. The operation is expressed as follows:
\begin{equation}
\mathbf{x}^{\prime} = \frac{\mathbf{x}^{\prime} - \alpha \cdot
\mathbb{E}[\mathbf{x}^{\prime}]}
{\sqrt{\textrm{Var}[\mathbf{x}^{\prime} - \alpha \cdot \mathbb{E}[\mathbf{x}^{\prime}]]
+ \epsilon}} \cdot \gamma + \beta
\end{equation}
where $\gamma$ and $\beta$ are learnable affine parameters that are similar to those used in other normalization techniques.

Afterward, The representations produced by the two streams,  ${X'}_{rgb}$ and ${X'}_{xyz}$, are finally fused  using a multi-head-cross attention layer ${MHSCA_{\boldsymbol{\Theta}}} : \mathbb{R}^{F'}\rightarrow \mathbb{R}^{F'}$. We note that cross-attention is only a varied form of self-attention and intuitive information fusion method in which attention from one distribution is used to highlight the extracted features in another distribution. The fusion here is carried out by measuring the similarity between the queries $\mathbf{q}$ (a linear projection  of ${X'}_{rgb}$) and the keys $\mathbf{k}$ (a linear projection  of ${X'}_{xyz}$) as well as using it to adjust the values vector $\mathbf{v}$ (a linear projection  of ${X'}_{xyz}$). Consequently,  in contrast to \textit{Equation} (3), ${MHCA_{\boldsymbol{\Theta}}}$ functions takes two inputs ${f_{}}({X'}_{xyz})$, and ${f_{}}({X'}_{rgb})$. The resulting vector is considered to be the new updated version of ${X'}_{xyz}$ that is used for the subsequent network embedding layer, We refer the reader to \cite{trans,cross-vit} for more details about the computation of cross-attention. To sum up, in the two-stream network, at each level a combination between the geometry and the color stream is applied to update the geometry representation with information about the corresponding color features.

After three consecutive (\textit{feature embedding}, \textit{MHSA} \textit{GraphNorm}, \textit{MHCA} ) blocks of layers, a \textbf{Points Embedding Aggregation} is applied on each patch  independently. Here a Max and Mean pooling operations  are applied on the features channel dimension. Afterward, the result of the two pooling is concatenated to one  vector, so each point cloud partition is with $M$ patch, is transformed into $M$ sequence of ${2F'}$-dimension vectors.

\subsubsection{Feature Aggregation and Quality Estimation}

In order to aggregate the representations obtained from each point cloud partition and capture the affinities between them. We used a \textbf{Patches Aggregation} method. This involved a multi-head self-attention layer~\cite{trans}, ~\cite{satsal} followed by max pooling to produce a vector representing for each partition. Finally, A shallow multi-layer perceptron was used to estimate the quality score, and the overall score was obtained by computing the mean of the sequence of partition scores.









\section{Experiments}

\subsection{Training and Implementation Details}

We trained our model end-to-end using the mean square error (MSE) as the loss function. The goal of the network is to create a mapping function between the input point clouds and the mean opinion score (MOS) quality. The loss function is defined as:

\begin{equation}
\mathcal{L} = \mathrm{MSE}\left(\mathrm{mean}\left(\sum_i{Out_{i}}\right), \mathcal{Y}\right),
\end{equation}

where $Out_{i}$ refers to each predicted partition score, and $\mathcal{Y}$ refers to the MOS.

The model was implemented in PyTorch and trained using the Adam optimizer \cite{Adam} with an initial learning rate of 0.0001 and a batch size of one. The number of epochs varied depending on the training folds, ranging from 80 to 200 in the ICIP dataset.

\subsection{Evaluation Protocol and Result Analysis}

To evaluate the effectiveness of our model, we conducted experiments on two publicly available benchmarks that use subjective scores and adopt different emerging compression schemes,  ICIP20 \cite{ICIP20} and PointXR \cite{pointxr}. ICIP20 includes 6 reference point clouds, each compressed using 5 levels and 90 degraded versions were derived through three types of compression. PointXR includes 5 point clouds, each compressed using G-PCC with octree coding for geometry compression and Lifting and RAHT for color compression, resulting in 45 degraded versions.
\begin{table}[t]
\small
\label{tab:tab1}
\caption{\label{tab:PXR}Results obtained by training the model on ICIP20 and testing it on PointXR}
\begin{center}
\begin{tabular}{ c c c  }
\hline
\textbf{Model} & \textbf{PLCC  $\uparrow$ } & \textbf{SROCC $\uparrow$ }   \\ 
\hline
 po2pointMSE &   0.887 & 0.978  \\ 
 \hline
 po2planeMSE & 0.855 & 0.942  \\
 \hline
 PSNRpo2pointMSE & \textbf{0.983} & \textbf { 0.978}  \\
 \hline
 PSNRpo2planeMSE & 0.972 & 0.950  \\
 \hline
 PointNet-DCCFR  & 0.981 &  {0.964}    \\
 \hline
PointGraph  & 0.967&  \textbf { 0.988 }    \\
 \hline
Our  & \textbf { 0.969}&   \textbf { 0.990 }   \\
 \hline

\end{tabular}
\end{center}
\vspace{-6mm}
\end{table}

We used a 6-fold cross-validation protocol to train and test our model on ICIP20, with 5 reference point clouds used for training and one for testing at each iteration. To evaluate the generalization ability of our method to predict the quality on unknown PCs, we also trained on ICIP20 and tested on PointXR using Pearson and Spearman correlations. Results are reported in Table~\ref{tab:icip_results}  and Table~\ref{tab:PXR}, with mean correlations calculated over all folds. We compared our method's performance to state-of-the-art methods using the same protocol.

Table~\ref{tab:icip_results} presents the results of our method on the ICIP20 dataset and compares them to state-of-the-art methods. Our results demonstrate a strong correlation with the subjective ground truth, showing a clear gap for both PLCC and SROCC when compared to most existing methods. Our proposed method outperforms all other methods with the highest SROCC score achieving a correlation equal to \textbf{0.976}, and sharing the highest PLCC score with the po2planeMSe method with a corelation equal to \textbf{0.959}. It's worth noting that our method even surpasses most of the full-reference methods. Notably, all listed methods in the table are full-reference except for PointNet-SSNR and PointNet-Graph, which are no-reference.
\begin{table}[t]
\small
\label{tab:tab2}
\caption{\label{tab:icip_results} Results obtained on ICIP20 dataset using 6-fold cross validation  
}
\begin{center}
\begin{tabular}{ c c c  }

\hline
\textbf{Model} & \textbf{PLCC  $\uparrow$ } & \textbf{SROCC $\uparrow$ }   \\ 
\hline
 po2point MSE & \textbf{ 0.946} & 0.934     \\ 
 \hline
 po2plane MSE & \textbf{0.959} & 0.951         \\
 \hline
 PSNR po2point MSE & 0.868 & 0.855         \\
 \hline
PSNR po2point HAU & 0.548 & 0.456 \\
\hline 
PSNR po2plane HAU  &0.580 & 0.547\\
\hline 
color Y MSE  &0.876  &0.892 \\
\hline
color Cb MSE &0.683& 0.694\\
\hline
color Cr MSE &0.594 &0.616 \\
\hline
color Y PSNR& 0.887& 0.892 \\
\hline
color Cb PSNR &0.693& 0.694\\
\hline 
color Cr PSNR &0.626& 0.617 \\
\hline
pl2plane AVG &0.922 &0.910 \\
\hline
pl2plane MSE &0.925& 0.912 \\
 \hline
 PCQM & 0.796 & 0.832       \\ 
\hline
GraphSim & 0.931 & 0.893     \\
\hline
 PointNet-SSNR &  {0.908} &  {0.955} \\
 \hline
 PointNet-DCCFR &  \textbf{ 0.947} &  \textbf{0.973} \\
 \hline
PointGraph &  \textbf{ 0.946}  & \textbf{0.973} \\
 \hline
Our &  \textbf{ 0.959}  & \textbf{0.976} \\
 \hline
\end{tabular}
\end{center}
\vspace{-6mm}
\end{table}


The cross-dataset evaluation results are presented in Table ~\ref{tab:PXR}, demonstrating high correlations achieved by our method and outperforming some of the compared methods. These results exhibit the generalization capability of our metric in predicting the quality of unknown point clouds, indicating the proposed method's consistent performance across different validation sets. As larger annotated datasets for point clouds quality become available in the future, further validation of our approach can be conducted.


\section{Conclusion}


In this paper, we introduce a novel no-reference quality metric for point clouds using a learning-based approach. Our network employs multi-head self and cross-attention mechanisms to capture the local semantic affinities and establish probabilistic connectivity between points. Since the perceptual quality degradation can occur on both geometry and color levels, we propose a two-stream architecture that processes color and geometry distortion in parallel, and interact dynamically at multiple levels of the network. Our method achieves state-of-the-art correlations on two benchmarks for point clouds quality assessment . 

\newpage
\label{sec:conclusion}

\bibliographystyle{IEEEbib}
\bibliography{refs}

\end{document}